\documentclass[letterpaper, 10 pt, conference]{IEEEtran}

\usepackage{cite}
\usepackage{graphicx}
\usepackage{framed}
\usepackage{subfig}
\usepackage{indentfirst}
\usepackage[]{algorithm2e}

\usepackage{amsmath, amssymb,amsthm}
\usepackage[font=small]{caption}

\usepackage{array}
\usepackage{makecell}
\usepackage{booktabs}
\usepackage{comment}
\usepackage{color}

\usepackage[pdftex, breaklinks]{hyperref}
\usepackage[capitalise]{cleveref}

\usepackage{multirow}
\usepackage{siunitx}
\usepackage{flushend}

\newcommand{\ls}[1]{\textbf{\textcolor{blue}{#1}}}

\IEEEoverridecommandlockouts                

\title{\LARGE \bf 
INTERACTION Dataset: \\An INTERnational, Adversarial and Cooperative moTION Dataset \\in Interactive Driving Scenarios with Semantic Maps}
\author{Wei Zhan$^{1}$, Liting Sun$^{1}$, Di Wang$^{2, \star}$, Haojie Shi$^{3, \star}$, Aubrey Clausse$^{4}$, Maximilian Naumann$^{5, \star}$,\\ Julius K\"ummerle$^{5}$, Hendrik K\"onigshof$^{5}$, Christoph Stiller$^{5}$, Arnaud de La Fortelle$^{4}$ and Masayoshi Tomizuka$^{1}$
\thanks{$^{1}$W. Zhan, L. Sun, and M. Tomizuka are with the Mechanical Systems Control (MSC) Laboratory, Department of Mechanical Engineering, University of California, Berkeley, CA 94720 USA. (e-mail: \tt\small wzhan@berkeley.edu).}
\thanks{$^{2}$D. Wang is with Xi'an Jiaotong University, Xi'an, P.R. China. }
\thanks{$^{3}$H. Shi is with Harbin Institute of Technology, Harbin, China. }
\thanks{$^{4}$A. Clausse and A. de La Fortelle are with MINES ParisTech, Paris, France. }
\thanks{$^{5}$M. Naumann, J. K\"ummerle, H. K\"onigshof and C. Stiller are with FZI Research Center for Information Technology and Karlsruhe Institute of Technology, Karlsruhe, Germany. }
\thanks{$\star$ The work was conducted during their visit to the MSC Lab at University of California, Berkeley.}
}

\begin{document}

\maketitle
\begin{abstract}
Interactive motion datasets of road participants are vital to the development of autonomous vehicles in both industry and academia. Research areas such as motion prediction, motion planning, representation learning, imitation learning, behavior modeling, behavior generation, and algorithm testing, require support from high-quality motion datasets containing interactive driving scenarios with different driving cultures. In this paper, we present an INTERnational, Adversarial and Cooperative moTION dataset (INTERACTION dataset) in interactive driving scenarios with semantic maps. 

Five features of the dataset are highlighted. 1) The interactive driving scenarios are diverse, including urban/highway/ramp merging and lane changes, roundabouts with yield/stop signs, signalized intersections, intersections with one/two/all-way stops, etc.
2) Motion data from different countries and different continents are collected so that driving preferences and styles in different cultures are naturally included.
3) The driving behavior is highly interactive and complex with adversarial and cooperative motions of various traffic participants. Highly complex behavior such as negotiations, aggressive/irrational decisions and traffic rule violations are densely contained in the dataset, while regular behavior can also be found from cautious car-following, stop, left/right/U-turn to rational lane-change and cycling and pedestrian crossing, etc.
4) The levels of criticality span wide, from regular safe operations to dangerous, near-collision maneuvers. Real collision, although relatively slight, is also included. 
5) Maps with complete semantic information are provided with physical layers, reference lines, lanelet connections and traffic rules. 

The data is recorded from drones and traffic cameras, and the processing pipelines for both are briefly described. Statistics of the dataset in terms of number of entities and interaction density are also provided, along with some utilization examples in the areas of motion prediction, imitation learning, decision-making and planing, representation learning, interaction extraction and social behavior generation. The dataset can be downloaded via \url{https://interaction-dataset.com}. 
\end{abstract}

	\begin{figure}[htbp]
	\begin{center}
	\includegraphics[width=8.5cm]{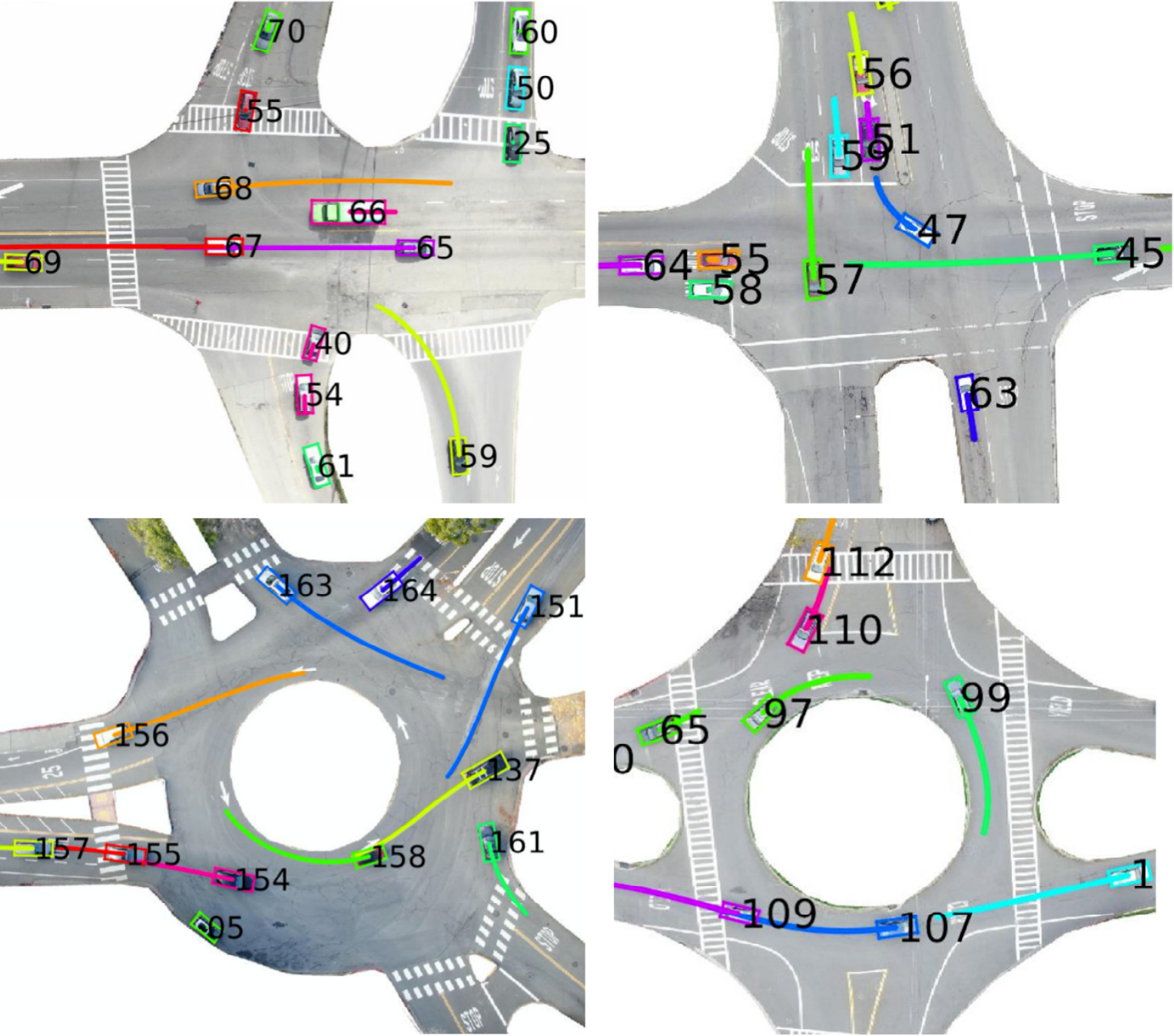}
	\caption{Examples of the detection and tracking results in highly interactive driving scenarios in the dataset.}
	\label{fig: illustration}
	\end{center}
	\end{figure}

\section{Introduction\label{sec: introduction}}
In order to enable fully autonomous driving in complex scenarios, comprehensive understanding and accurate prediction of the behavior and motion of other road users are required. Moreover, autonomous vehicles need to behave like vehicles with human drivers to make themselves more predictable to others and thus, facilitate cooperation. These are two of the major challenges in the field of autonomous driving. To overcome these challenges, considerable amount of research efforts have been devoted to: i) predicting the future intention and motion of other road users \cite{lefevre_survey_2014, rudenko2019human, zhan_probabilistic_iv2018}, ii) modeling and analyzing driving behavior \cite{okuda_modeling_2013, driggs-campbell_integrating_2017}, iii) clustering the motion and finding representation of the motion primitives \cite{lin_moha_2018, wang_understanding_2018}, iv) cloning and imitating human and expert behavior \cite{kelly_hg-dagger_2019, rhinehart_deep_2018}, and v) generating human-like and social behavior and motion \cite{sun_courteous_2018, guo_toward_2018, naumann2018comfortable}.

All the aforementioned research areas require interactive vehicle motion data from real-world driving scenarios, which is the most fundamental and indispensable asset. NGSIM dataset \cite{alexiadis_next_2004} is the most popular one used in the aforementioned areas, such as prediction \cite{sun_probabilistic_2018, zhan_towards_2018, altche_lstm_itsc2017}, behavior modeling \cite{driggs-campbell_integrating_2017}, social behavior generation and planning \cite{sun_courteous_2018}, and representation learning \cite{lin_moha_2018}, since it is publicly available with decent scale and quality. The recently released highD dataset \cite{krajewski_highd_2018} also greatly assists behavior-related research such as prediction \cite{messaoud2019relational}. Public motion datasets such as NGSIM and highD facilitated, but also restricted behavior-related research due to limited diversity, complexity and criticality of the scenarios and behavior. Also, the importance of map information and completeness of interaction entities were under-addressed in most of the existing datasets. However, these missing points are crucial for behavior-related research, which will be discussed in the following. 

\textit{1) Diversity of interactive driving scenarios:} Recent behavior-related research using public datasets was mostly restricted to highway scenarios due to the data availability. There are many more highly interactive driving scenarios to explore, such as roundabouts with yield/stop signs, (unsignalized) one/two/all-way stop intersections (shown in \cref{fig: illustration}), signalized intersections with unprotected left turn, zipper merge in cities, etc.

\textit{2) International diving cultures:} Most of the existing datasets only contain driving data in one specific country. However, driving cultures in different countries and different continents can be distinct for very similar scenarios. Without motion data in similar scenarios from different countries, it is not possible to incorporate the impact of driving cultures in different countries, such as driving styles, preferences, risk tolerance, understanding of traffic rules, etc., for behavior modeling and analysis as well as the design of adaptive prediction and planning algorithms in different countries.

\textit{3) Complexity of the scenarios and behavior:} Most of the scenarios in the existing public datasets are relatively simple and structured with explicit right-of-way. The behavior of the drivers is only occasionally impacted by others. There is very little social pressure (such as several vehicles waiting behind and even honking) on the drivers, so that their behavior is cautious without aggressive and irrational decisions. A motion dataset with much more complex and interactive behavior and scenarios is expected to facilitate the research tackling real and challenging problems.

\textit{4) Criticality of the situations:} Critical situations (such as near-collision cases) are much more challenging and valuable than others for behavior-related research areas. For instance, \cite{zhan_towards_2018} proposed a fatality-aware prediction benchmark emphasizing prediction inaccuracies in critical situations. However, critical situations are too sparse in existing motion datasets, and can hardly be identified. Therefore, a motion dataset with denser critical situations is necessary to facilitate the research efforts on those difficult problems.

\textit{5) Map information:} Map information with references and semantics such as lanelet connections and traffic rules, are crucial for behavior-related research areas such as motion planning and prediction. It provides key information on input (features), such as route and goal point \cite{rhinehart_deep_2018}, distance to the merging point \cite{sun_probabilistic_2018, zhan_towards_2018}, lateral position within the lane \cite{sun_courteous_2018}, etc., and makes the algorithms generalizable to other scenarios. Such semantic maps are currently missing for most of the existing public motion datasets.

\textit{6) Completeness of interaction entities:} In order to accurately model, predict and imitate the interactive vehicle behavior, it is crucial to provide motions of all surrounding entities which may impact their behavior in the dataset. This requirement was often overlooked when using motion data collected by onboard sensors due to occlusions and limited field of view of the sensors. Although existing motion datasets collected from onboard sensors contain data collected from a wide range of areas for long time periods, complete and meaningful interaction pairs are relatively sparse.

In this paper, we will emphasize all the aforementioned aspects to construct an international motion dataset collected by drones and traffic cameras. 
\begin{itemize}
    \item \textit{Diverse and international:} It contains a variety of highly interactive driving scenarios from different countries, such as roundabouts, signalized/unsignalized intersections, as well as highway/urban merging and lane change. 
    \item \textit{Complex and critical:} Part of the scenarios are relatively unstructured with inexplicit right-of-way. The driving behavior in the dataset are highly impacted by other drivers, whose behavior can be aggressive or irrational due to the social pressure. Near-collision or slight-collision scenes are contained in the dataset to facilitate the research for critical situations.
    \item \textit{Semantic map and complete information:} HD maps with semantics are provided to generated key features in the context. Motions of all entities which may influence the driving behavior are included in the dataset. 
\end{itemize}
The proposed dataset can significantly facilitate behavior-related research such as motion prediction, imitation learning, decision-making and planning, representation learning, interaction extraction and social behavior generation. Results from exemplar methods in all these areas are provided utilizing the proposed dataset.

\section{Related Work\label{sec: relatedwork}}

\begin{table*}[tbp]
	\caption{Comparison with existing motion datasets}\centering{}%
	\begin{tabular}{c|ccccccc}
		& \makecell{highly interactive\\ scenarios} & \makecell{complexity of\\ scenarios} & \makecell{density of\\ aggressive\\ behavior} & \makecell{near-collision\\ situations\\ and collisions} & \makecell{HD maps \\with semantics} & \makecell{completeness of\\ interaction entities\\ \& viewpoint}  \\ \hline \hline
		NGSIM \cite{alexiadis_next_2004} & \makecell{ramp merging,\\ (double) lane change} & \makecell{structured roads,\\ explicit right-of-way} & low & \makecell{very few\\ near-collision} & no & \makecell{yes, bird's-eye-view\\ from a building} \\ \hline
		highD \cite{krajewski_highd_2018} & \makecell{lane change} & \makecell{structured roads,\\ explicit right-of-way} & low & \makecell{very few\\ near-collision} & no & \makecell{yes, bird's-eye-view\\ from a drone} \\ \hline
		Argoverse \cite{chang_argoverse_2019} & \makecell{unsignalized intersections,\\pedestrian crossing} & \makecell{unstructured roads,\\ inexplicit right-of-way} & low & \makecell{no} & \makecell{yes, \\but partially} & \makecell{only for the ego\\ data-collection vehicle} \\ \hline
		INTERACTION & \makecell{roundabouts, ramp merging,\\ double lane change\\ unsignalized intersections} & \makecell{unstructured roads,\\ inexplicit right-of-way} & high & yes & yes & \makecell{yes, bird's-eye-view\\ from a drone}%
	\end{tabular}%
\label{table: 1}
\end{table*}

\subsection{Datasets from Bird's Eye View\label{subsec: bird}}
As mentioned in Section \ref{sec: introduction}, NGSIM dataset \cite{alexiadis_next_2004} is the most popular vehicle motion dataset among the behavior-related research communities. The raw data was collected by cameras mounted on buildings and processed automatically \cite{kim2005machine}. The accuracy of the dataset is mostly acceptable. However, there may be steady errors, and the image projection can significantly enlarge the size of the vehicles. Researchers proposed methods \cite{coifman_critical_2017} to rectify the errors, but it can only improve the quality of a small part of the dataset. In view of the problems in NGSIM, highD dataset \cite{krajewski_highd_2018} was constructed by using a drone with more accurate vehicle motions and larger amount of high way driving data than NGSIM. Other datasets \cite{yang_top-view_2019, robicquet_learning_2016} from bird's eye view are more focused on pedestrian behavior without strong vehicle interactions.

The driving scenarios presented in NGSIM and highD are quite limited. NGSIM contains highway driving (including ramp merging and double lane change) and signalized intersection scenarios. In fact, signalized intersections are mostly controlled by the traffic lights and interactions are very rare and slight. A small amount of lane changes are interactive, but most of them are neither interactive nor critical. Ramp merging and double lane change can be highly interactive when the traffic is relatively dense, but the amount of interaction is still relatively limited in NGSIM. HighD only contains highway driving scenarios with car following and lane change. Urban scenarios which contain densely and highly interactive behavior, such as roundabouts and unsignalized intersections are not included in either of the two public datasets of vehicle motions.

\subsection{Datasets from Onboard Sensors\label{subsec: onboard}}

In addition to the bird's-eye-view motion datasets, two types of onboard-sensor-based ones are also publicly available. One includes motion data of surrounding entities from onboard LiDARs and front-view cameras, such as Argoverse \cite{chang_argoverse_2019} and HDD dataset \cite{ramanishka_toward_2018}. The other only contains motions of many data-collection vehicles from onboard GPS, such as 100-car study \cite{neale_overview_2005}. 

There are two major advantages for datasets from onboard sensors. One is that a variety of driving scenarios with relatively long data recording time are usually included in those datasets, such as urban driving at signalized/unsignalized intersections and highway driving with ramp merging, etc. The other is that the occlusions of LiDARs and cameras are recorded so that the actual occlusions from perspective of the ego vehicle can be partially recovered.

Completeness of interaction entities is a major problem when using datasets from onboard sensors for behavior-related research. For motion datasets with GPS-based fleets, it is hard to determine whether the vehicles in an "interactive" motion segment was actually interacting with each other since there is no motion recording of other surrounding vehicles (or even pedestrians) without GPS devices installed. For motion datasets constructed from onboard LiDARs and cameras, it is hard to guarantee that all the surrounding objects impacting the behavior of other vehicles are included in the dataset when predicting the motions of others. Therefore, complete interactions are relatively sparse in such kind of datasets. If the sensors cannot cover the full field of view, it will be even impossible to guarantee the completeness of information for the surrounding entities of the ego data collection vehicle.

Also, the data collected in a large area may lead to very few repetitions at the same location. It is hard to learn multi-modal driving behavior for prediction or planning since only one sequence of motions can be found with similar features at the same location. 

Map information is also missing in most of the motion datasets. To the best of our knowledge, Argoverse is the only motion dataset providing relatively rich map information. Physical layer (locations of curbs, road markings, etc.) is contained and semantic information (lane bounds and turn directions, etc.) required by prediction and planning is partially included. 

\cref{table: 1} provides a comparison of the three most useful public vehicle motion datasets as well as the one presented in this article. The proposed dataset contains much more diverse, complex and critical scenarios and vehicle motions comparing to the other three. In addition, HD maps with full semantic information are provided, and the completeness of interaction entities is superior to datasets from onboard sensors.

	\begin{figure*}
	\begin{center}
	\includegraphics[width=16.5cm]{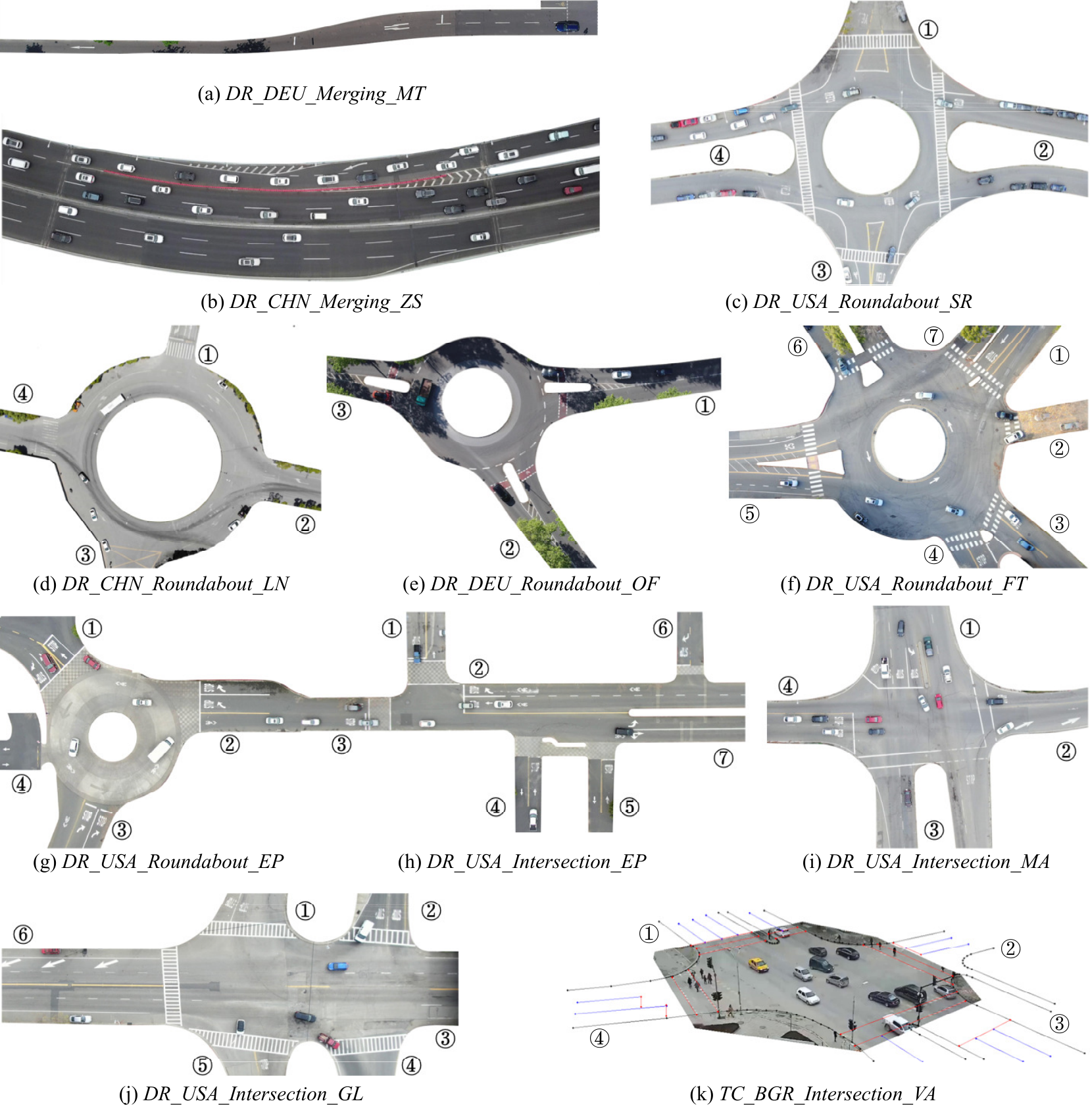}
	\caption{A variety of highly interactive driving scenarios recorded by drones in the dataset, including: (a) urban merging, (b) highway ramp merging and lane change, (c)-(g) five roundabouts, and (h)-(j) unsignalized intersections, and (k) unprotected left turn at a signalized intersection.}
	\label{fig: scenarios}
	\end{center}
	\end{figure*}

\section{Features of the Dataset\label{sec: scenario}}

In this section, we will illustrate the features of the proposed dataset by highlighting the diversity, internationality, complexity, criticality, and semantic map.

\subsection{Diversity\label{subsec: diversity}}

\cref{fig: scenarios} illustrates a variety of highly interactive driving scenarios from traffic cameras and drones in our dataset, including zipper merging in a city (\cref{fig: scenarios} (a)), ramp merging and lane change on a highway (\cref{fig: scenarios} (b)), five roundabouts with yield and stop signs (\cref{fig: scenarios} (c) - (g)), several unsignalized intersections with one/two/all-way stops (\cref{fig: scenarios} (h) - (j)), and unprotected left turn at a signalized intersection (\cref{fig: scenarios} (k)). In \cref{fig: scenarios}, the first two letters of the names represent the sources of the data (drone as \textit{DR} and traffic camera as \textit{TC}), while next three letters represent the corresponding country and the last two represent the scenario code in the dataset. The numbers in circles denote the branch ID for each scenario.

\cref{fig: scenarios} (b) contains several subscenarios. The subscenario with the upper two lanes (that merge into one finally) is a zipper merging which is similar to the urban counterpart in \cref{fig: scenarios} (a), where vehicles strongly interact with each other. It is also a ramp for the middle two lanes. The subscenario with the lower three lanes (that merge into two finally) is a forced merging and vehicles have to change their lanes.

The roundabout in \cref{fig: scenarios} (f) is an extremely busy 7-way roundabout with one ``yield'' branch and six ``stop'' branches. Lots of vehicles enter the roundabout at the same time with intensive interactions and relatively high speeds. The branches of the roundabouts in \cref{fig: scenarios} (c)-(e) are controlled by yield signs, while all branches of the roundabout in \cref{fig: scenarios} (g) are controlled by stop signs. 

\Cref{fig: scenarios} (i) shows an extremely busy all-way-stop intersection with 9 lanes controlled by stop signs. Multiple vehicles are interactively inching to compete. The scenario shown in \cref{fig: scenarios} (j) contains three branches (Branch 1, 2, 5) controlled by stop signs, while vehicles from Branch 3 and 6 have the right-of-way (RoW). Lots of vehicles are entering the intersections from all branches (except Branch 4), and vehicles holding RoW on the straight road are with relatively high speed. A busy all-way-stop T-intersection is shown in \cref{fig: scenarios} (h), while three other branches (Branch 4-6) are also controlled by stop signs.

\subsection{Internationality\label{subsec: Internationality}}

The motion data was collected from three continents (North America, Asia and Europe). Motion data collected by drones are from four countries, namely, the US, China, Germany and Bulgaria, as indicated in the names of the scenarios (\textit{USA/CHN/DEU/BGR}). Vehicles in all these countries are driven on the right-hand side of the road. However, driving culture in these countries is with remarkable distinctions.

We provide motion data from three roundabouts with similar traffic rules, namely, \textit{SR} from the US, \textit{OF} from Germany and \textit{LN} from China. All the three roundabouts do not have stop signs, and the nominal traffic rule is that the vehicles entering the roundabout should yield the ones which is already in the roundabout.

We also provide motion data from two zipper merging scenarios, those are, \textit{MT} from Germany and \textit{ZS} from China (the upper two lanes in \cref{fig: scenarios} (b)). Although \textit{MT} is urban road and \textit{ZS} is the entrance of highway, the ``zipper'' rule remains the same, and the speeds are similar when the traffic is heavy.

\subsection{Complexity\label{subsec: complexity}}

In addition to regular driving behavior such as car-following, lane change, stop and left/right/U-turn, our dataset emphasizes highly interactive and complex driving behavior with cooperative and adversarial motions of the vehicles. By carefully choosing the locations and corresponding rush hours for the data collection, we were able to gather large amounts of strong interactions within relative short period of time. Strongly interactive pairs of vehicles can even appear every few seconds from time to time for scenarios such as the ramp in \textit{ZS}, the entrance branches in \textit{FT}, the all-way-stop intersections in \textit{EP} and \textit{MA} as well as the two-way-stop intersection in \textit{GL}. 

	\begin{figure}[htbp]
	\begin{center}
	\includegraphics[width=8.5cm]{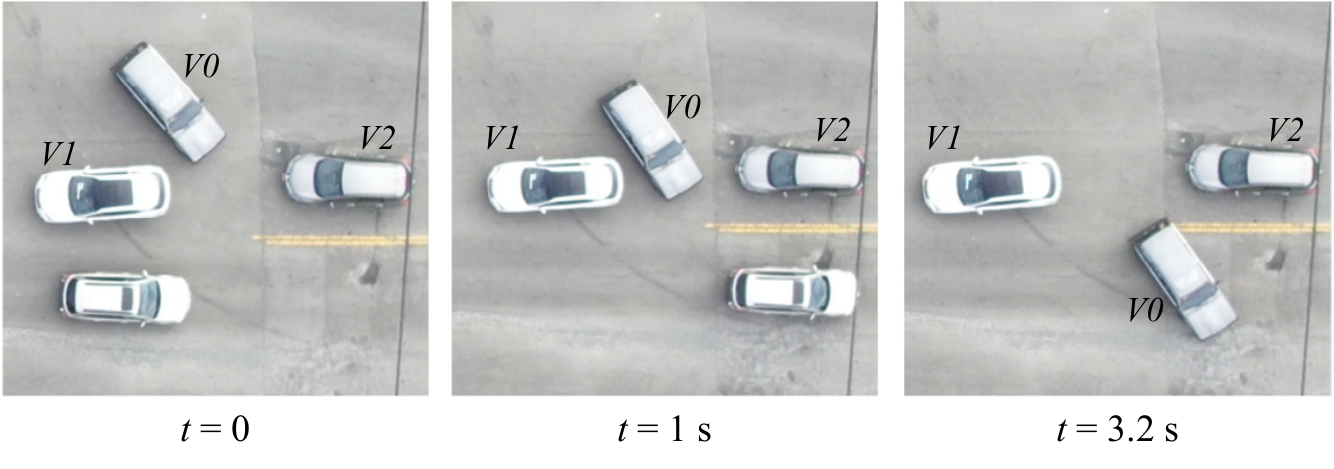}
	\caption{A sequence of images of a dangerous insertion in \textit{GL} in the proposed dataset.}
	\label{fig: dangerous-behavior}
	\end{center}
	\end{figure}

Also, scenarios in \textit{FT} and \textit{GL} are relatively unstructured since there is no explicit lane restrictions in the roundabout or intersections. Vehicles can exploit the space to achieve their goals, sometimes showing irrational and highly dangerous behavior. For instance, \cref{fig: dangerous-behavior} shows a dangerous insertion of \textit{V0} between two vehicles (\textit{V1} and \textit{V2}) stopping and making left turns from Branch 3 to Branch 5 in \textit{GL} (refer to \cref{fig: scenarios}). The driver of \textit{V0} intended to drive from Branch 1 to Branch 4 but there was no explicit road structure for the driver.
	
    \begin{figure}[htbp]
	\begin{center}
	\includegraphics[width=8.5cm]{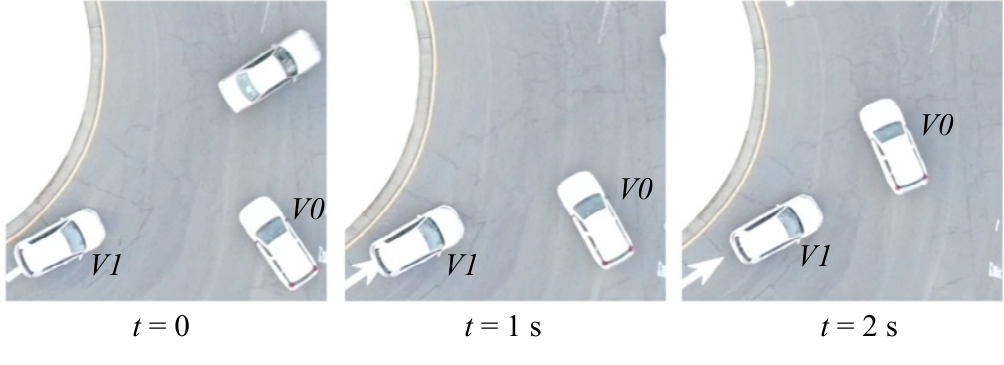}
	\caption{A sequence of images of a violation for the right-of-way in a roundabout in the proposed dataset.}
	\label{fig: violation-right-of-way}
	\end{center}
	\end{figure}

Moreover, aggressive or irrational behavior can often be found due to inexplicit nominal or practical RoW. Vehicles may arrive at the stop bars almost at the same time and drivers may negotiate with each other by inching or even accelerating in \textit{MA} and \textit{EP}. The traffic in \textit{FT} and \textit{GL} can be very busy and it may take even minutes for the vehicle without nominal RoW to enter and pass, making the driver impatient. Also, there may be a queue of vehicles waiting behind and even honking to put social pressures to the one in the front of queue. Although there are explicit traffic rules on who goes first for roundabouts or 2-way-stop intersections, vehicles without nominal RoW may be aggressive, and vehicles with nominal RoW are mostly aware of such potential violations and are ready to react. For example, \textit{V0} in \cref{fig: violation-right-of-way} was entering the roundabout in \textit{FT} from Branch 3, while \textit{V1} was in the roundabout holding the RoW. However, \textit{V0} violated the rule and forced \textit{V1} to stop and yield.

Those factors significantly increase the complexity of the motions in the dataset and bring forward lots of challenging but valuable research topics for the community.

\subsection{Criticality\label{subsec: criticality}}
	
As discussed in \cref{subsec: complexity}, vehicles holding the nominal RoW (in the roundabout of \textit{FT} or on the straight road of \textit{GL}) may often encounter slight violations from vehicles without nominal RoW (entering the roundabout or intersection from branches controlled by stop signs). Moreover, the vehicles holding the RoW may have relatively high speed (40 km/h or even higher). Therefore, critical situations can be observed in the dataset where time-to-collision-point (TTCP) can be extremely low. A slight collision can even be found in the dataset. 

	\begin{figure}[htbp]
	\begin{center}
	\includegraphics[width=8.5cm]{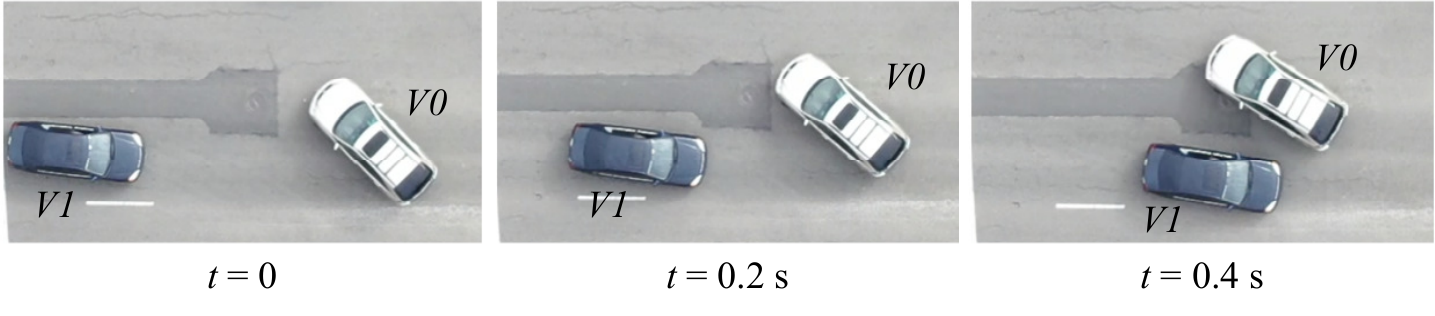}
	\caption{A sequence of images of a near-collision case in the proposed dataset.}
	\label{fig: fast-near-collision}
	\end{center}
	\end{figure}

\cref{fig: fast-near-collision} shows a near-collision case in \textit{GL}. \textit{V0} was making a left turn from Branch 5 (with a stop sign) to Branch 6, while \textit{V1} (with the RoW) was going straight forward from Branch 6 to Branch 3 with a relatively high speed. \textit{V1} had to execute emergency swerve to avoid the collision with \textit{V0}, which was very dangerous.

	\begin{figure}[htbp]
	\begin{center}
	\includegraphics[width=8.5cm]{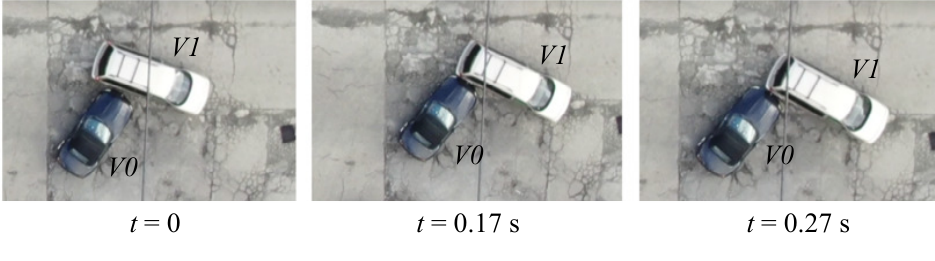}
	\caption{A sequence of images of a slight collision in the proposed dataset.}
	\label{fig: collision}
	\end{center}
	\end{figure}

Besides the critical, near-collision cases, a slight collision shown in \cref{fig: collision} can also be found in the dataset in \textit{GL}. \textit{V0} was making a right turn from Branch 5 (with a stop sign) to Branch 3, while \textit{V1} (with the RoW) was making a right turn from Branch 6 to Branch 4. In this situation, the driver of \textit{V0} might have predicted that \textit{V1} was going straight to Branch 3, so that \textit{V0} could accelerate in advance.

\subsection{Semantic Map\label{subsec: Semantic}}

Map information is crucial for behavior-related research areas. The information required is twofold. The basic requirement is the physical layer containing a set of points or curves representing curbs, road markings (lane markings, stop bars, etc.) and other key features. In addition to the physical layer, semantic information is also necessary, which includes but is not limited to, 1) reference paths, 2) lanelets as well as their connections and turn directions, 3) traffic rules and RoW associated, etc. Moreover, such information needs to be organized with consistent format and toolkit to facilitate the users when utilizing the map. All the aforementioned requirements are met in our dataset, and more detailed information on map construction can be found in \cref{sec: map}.

\section{Construction of Motion Data and Maps\label{sec: processing}}

In this section, we will discuss the pipeline for constructing the motion data from both drones and traffic cameras, as well as the corresponding semantic maps.

\subsection{Motions from Drone Data\label{subsec: drone construction}}

We used drones such as DJI Mavic 2 and DJI Phantom 4 to collect the raw video data. The raw videos were 4K (3840x2160) by \SI{30}{Hz}. We downsampled the video to \SI{10}{Hz} and process the data. The processed results are partially illustrated in \cref{fig: illustration}. The bounding boxes are very accurate and the paths are smooth after going through out processing pipeline with the following three steps.

\begin{itemize}
    \item \textit{Video stabilization and alignment:} Due to gradual or sudden drift and rotation of drones, the collected videos need to be stabilized via video stabilization algorithms with transformation estimator. Also, similarity transformation is applied to project all the frames to the first one and aligned with the map.
    \item \textit{Detection:} In order to obtain accurate bounding boxes of the moving obstacles, Faster R-CNN \cite{ren_faster_2015} is applied. The boxes are highly accurate, and very few inaccurate detections are rectified manually. 
    \item \textit{Data association, tracking and smoothing:} Kalman filter is applied for data association and tracking. To obtain smooth motions of the vehicles, a Rauch-Tung-Striebel (RTS) smoother \cite{Simon:2006:OSE:1146304} is also incorporated.
\end{itemize}

\subsection{Motions from Traffic Camera Data\label{traffic camera construction}}

The data processing pipeline for motions from traffic camera data mainly contains the following steps, and more details, including the camera parameter estimation, can be found in \cite{ClausseIV2019}.

\begin{table*}[t]
	\centering
	\caption{Summary of the dataset.\label{tab:dataset_stat_summary}}
	\begin{tabular}{|c|c|c|c|c|c|}
		\hline
		Scenarios & Locations & Video length (min) & \makecell{number \\of vehicles} & \makecell{Total video \\ length (min)} & \makecell{Total number\\ of vehicles} \\
		\hline
		\multirow{5}{*}{roundabout} & USA\_Roundabout\_SR & 40.90 & 965 & \multirow{5}{*}{365.1} & \multirow{5}{*}{10479}\\
		\cline{2-4}
		& CHN\_Roundabout\_LN & 24.24 & 227 & &\\
		\cline{2-4}
		& DEU\_Roundabout\_OF & 55.04 & 1083 & &\\
		\cline{2-4}
		& USA\_Roundabout\_FT & 207.62 & 7496 & &\\
		\cline{2-4}
		& USA\_Roundabout\_EP & 37.30 & 708 & &\\
		\hline
		\multirow{3}{*}{unsignalized intersection} & USA\_Intersection\_EP & 66.53 & 1367 & \multirow{3}{*}{433.33} & \multirow{3}{*}{14867}\\
		\cline{2-4}
		& USA\_Intersection\_MA & 107.37 & 2982 & & \\
		\cline{2-4}
		& USA\_Intersection\_GL & 259.43 & 10518 & & \\
		\hline
		\multirow{2}{*}{\makecell{merging and\\ lane change}} & DEU\_Merging\_MT & 37.93 & 574 & \multirow{2}{*}{132.55} & \multirow{2}{*}{10933}\\
		\cline{2-4}
		& CHN\_Merging\_ZS & 94.62 & 10359 & & \\
		\hline
		\multirow{1}{*}{signalized intersection} & TC\_Intersection\_VA & 60 & 3775 & 60 & 3775\\
		\hline
	\end{tabular}	
\end{table*}

\begin{itemize}
	\item \textit{Detection:} To detect vehicles and pedestrians in each frame, we use a state-of-the-art object detector \cite{He2017MaskR}, which provides detections with 2D bounding box, instance mask and instance type.
	\item \textit{Data association:} Detections are grouped into tracks using a combination of an Intersection-over-Union \cite{1517Bochinski2017} tracker which associates detections with high mask overlap in successive frames, and a visual tracker \cite{Lukezic_IJCV2018} to compensate for miss detections.
	\item \textit{Tracking and smoothing:} Once detections are grouped into tracks, trajectories on the ground plane are estimated using a RTS smoother. For the observation model, we use a pin-hole camera model \cite{Brown71close-rangecamera}. This allows to incorporate measurements and uncertainty directly in pixels, capturing the uncertainty due to the resolution, position and orientation of the camera. For vehicles, the RTS smoother uses a bicycle model \cite{polack_kinematic_iv2017} as process model, allowing to capture the kinematics constraints of vehicles.
\end{itemize}

\subsection{Construction of the High Definition Maps\label{sec: map}}

As public roads are structured environments, the particular road layout of a certain area strongly affects the motion of all traffic participants.
The structure for vehicles mostly starts by subdividing the road into lanes, and later combining them to create junctions, roundabouts, on ramps and so on.
Further, movement within this structured area is guided by traffic rules, such as speed limits or prioritizing one road over another.
In order to model such coherence, simply mapping center-lines of all lanes is not sufficient anymore.

\begin{figure}
	\centering
	\includegraphics[trim={0cm 0cm 2cm 0cm},clip, width=\linewidth]{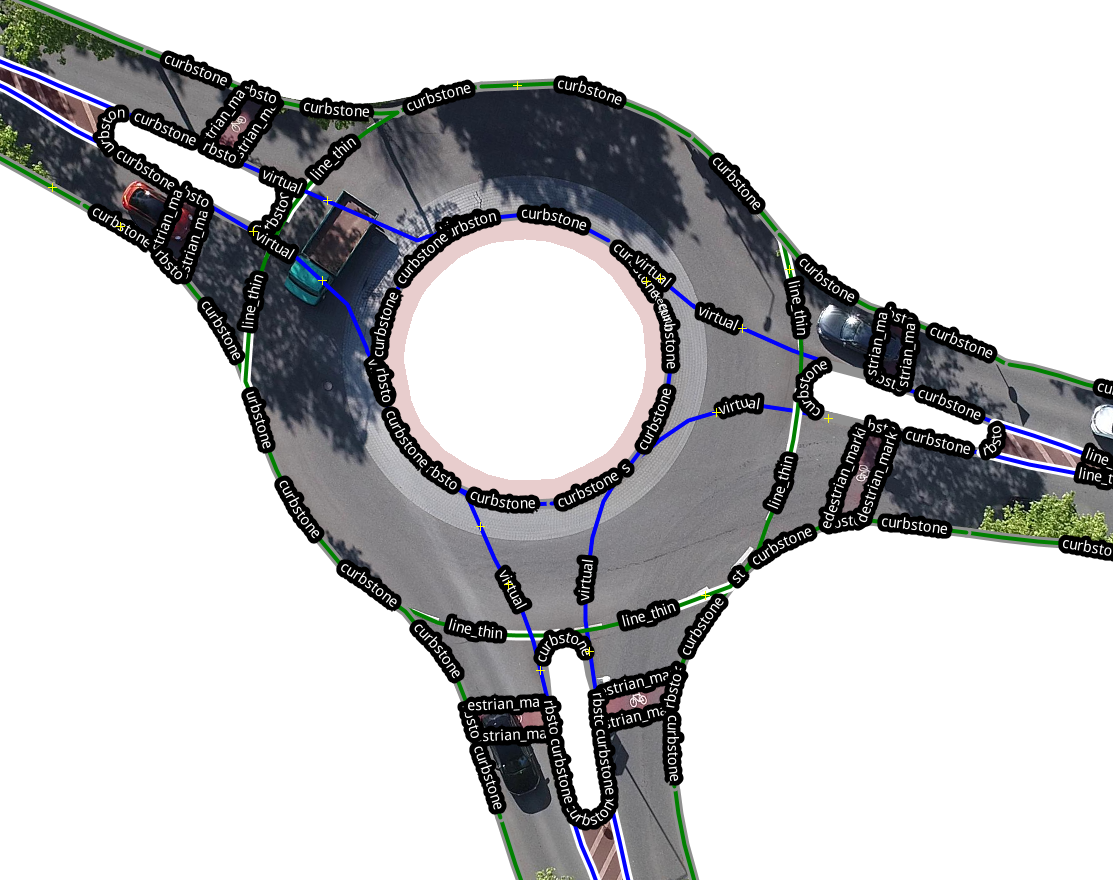}	
	\caption{An exemplary physical layer of a lanelet2 map \cite{poggenhans_lanelet2:_2018}.}
	\label{fig:lanelet2}
\end{figure}
 
Thus, in order to allow for a thorough analysis of the recorded trajectories, we provide centimeter-accurate high definition maps in the lanelet2 format \cite{poggenhans_lanelet2:_2018}.
Within lanelet2, the physical layer of the road network, such as road borders, lane markings and traffic signs is stored.
An exemplary physical layer is visualized in Figure \ref{fig:lanelet2}.
From this layer, atomic lane elements, called \textit{lanelets}, are created.
They describe the course of the lane and form the basis for so called regulatory elements, which determine traffic regulations such as the right of way or the speed limit.

When used alongside the recorded trajectories, these lanelet2 maps facilitate the reasoning about why some vehicles decelerate while approaching a junction, or why others do not, depending on the right of way but also on the presence of other traffic participants that potentially interact.

\section{Statistics of the Dataset\label{sec: statistics}}

\subsection{Scenarios and Vehicle Density}
The dataset contains motion data collected in four categories of scenarios: roundabout, unsignalized intersection, signalized intersection, merging and lane change, as shown in \cref{fig: scenarios}. A detailed summary of the dataset is listed in \cref{tab:dataset_stat_summary}. In the roundabout scenarios, 10479 trajectories of vehicles from five different locations were recorded for around 365 minutes. Similarly, in the unsignalized intersection scenarios, three locations were included and 14867 trajectories were collected for around 433 minutes. In the merging and lane change scenarios, 10933 trajectories were recorded at two locations for around 133 minutes. Finally, one location was selected for the signalized intersection, which provided 3775 trajectories for around 60 minutes.

\subsection{Metrics for Interactive Behavior Identification}
\begin{figure}[h!]
	\centering
	\includegraphics[width=0.45\textwidth]{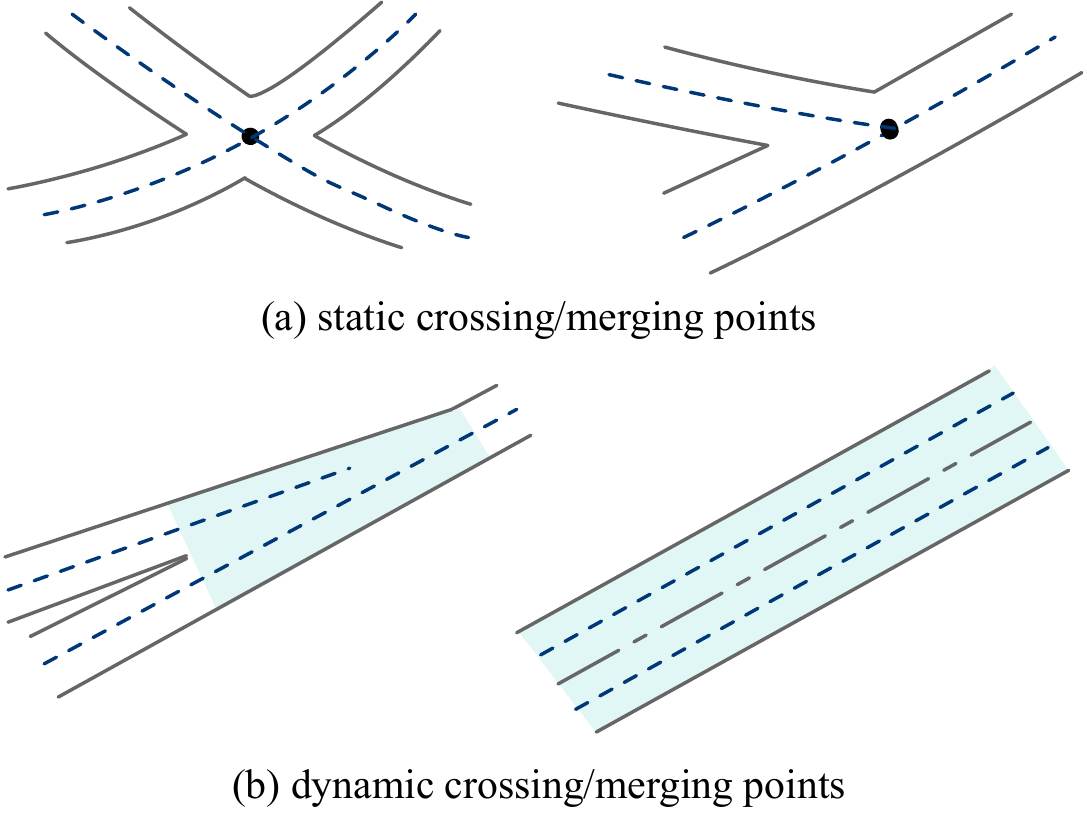}
	\caption{Geometry of different interactive paths. In (a), the crossing/merging points between two paths are static and fixed, while in (b), the crossing/merging points are dynamic.}
	\label{fig:crossing_point}
\end{figure}

To represent the density of the interactive behavior of the proposed dataset, we use the metric - number of interaction pairs per vehicle (IPV) as in proposed in \cite{zhan_constructing_iros2019}. To calculate the IPV, a set of rules were proposed in \cite{zhan_constructing_iros2019} to extract the interactive behavior under different spatial representations of vehicle paths. The set of rules and metric are briefly reviewed below.
\begin{enumerate}
	\item Minimum time-to-conflict-point difference ($\triangle TTCP_{\min}$): $\triangle TTCP_{\min}$ is a metric to describe the relative states of two moving vehicles in a scenario where the paths of the two vehicles share a conflict point but without any forced stop. As shown in \cref{fig:crossing_point}, such vehicle paths include two categories: (1) paths with static crossing or merging points such as intersections (\cref{fig:crossing_point} (a)-(b)), and (2) paths with dynamic crossing or merging points such as ramping and lane-changing, as shown in \cref{fig:crossing_point} (c)-(d). In such scenarios, merging can happen anywhere in the shaded area. We define $\triangle TTCP_{\min}$ as
	\begin{IEEEeqnarray}{rCl}
	\triangle TTCP_{\min}&=&\min_{t\in[T_{\text{start}}, T_{\text{end}}]} \triangle TTCP^t\nonumber\\ &=& \min_{t\in[T_{\text{start}}, T_{\text{end}}]} (TTCP_1^t-TTCP_2^t)\label{eq:TTC}
	\end{IEEEeqnarray}
where $TTCP_i^t = {\triangle d_i^t}/{v_i^t}, i=1,2$ is the traveling time to the conflict point of each vehicle in the interactive pairs. $v_i^t$ and $\triangle d_i^t$ are, respectively, the speed of the $i$-th vehicle and its distance to the conflict point along the path at time $t$. For the scenarios with dynamic merging points, we use the actual merging points of the vehicle trajectories as the conflict points. In (\ref{eq:TTC}), $T_{\text{start}}$ and $T_{\text{end}}$ are set to be long enough to cover the interaction period between vehicles. If $TTCP_{\min}\le$\SI{3}{s}, then it is defined that interaction exists. 

\item Waiting Period (WP): WP is a metric for vehicles with forced stops along their paths. In \cite{zhan_constructing_iros2019}, the default waiting period at stops was set as \SI{3}{s}, and the behavior deviation from the default one was used as an indicator of the interactivity, i.e., interaction exists when $\text{WP}>$\SI{3}{s}.
\end{enumerate}

\subsection{Distribution of Interactivity}
\begin{figure*}
	\centering
	\includegraphics[width=\textwidth]{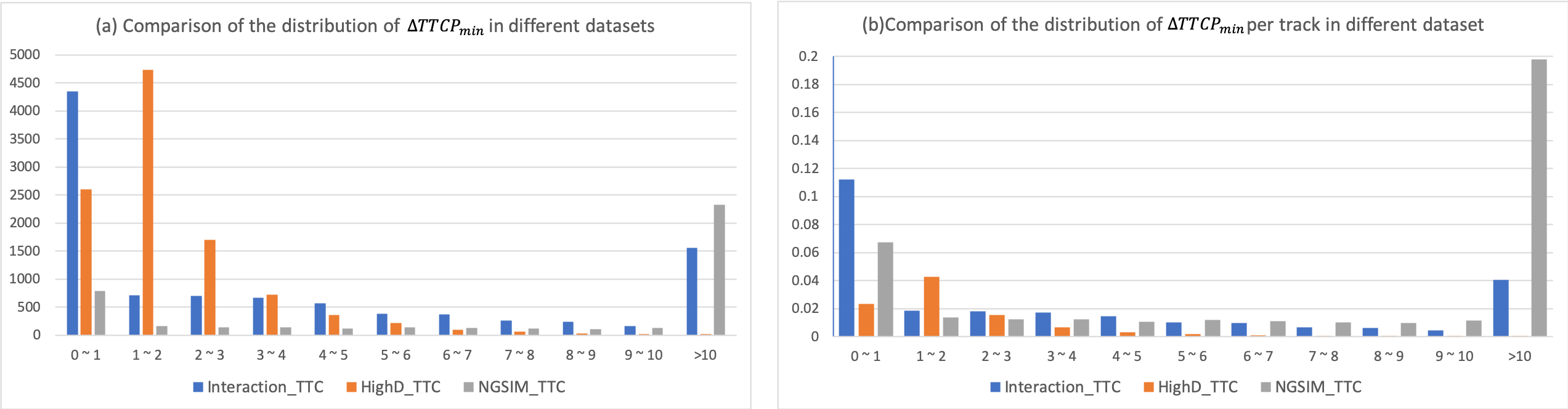}
	\caption{Distribution of the $\triangle TTCP_{\min}$ in three vehicle motion datasets: the proposed INTERACTION dataset, the HighD dataset and the NGSIM dataset.}
	\label{fig:distribution_TTC_across_dataset}
\end{figure*}

Based on the set of rules, there are 13375 interactive pairs of vehicles in the proposed dataset. We compare the interactivity among three datasets: the proposed INTERACTION dataset, the highD dataset, and the NGSIM dataset. Results are shown in \cref{fig:distribution_TTC_across_dataset}, where the x-axis represents the length of $\triangle TTCP_{\min}$ in seconds, and the y-axis are the number of vehicles (\cref{fig:distribution_TTC_across_dataset} (a)) and the density of vehicles\footnote[1]{The density is given by:\\ $\text{density}=\dfrac{\text{number of vehicles with particular } \triangle TTCP_{\min}}{\text{total number of vehicles in the dataset}}$.} (\cref{fig:distribution_TTC_across_dataset} (b)), respectively. We can see that the INTERACTION dataset contains more intensive interactions with $\triangle TTCP_{\min}\le1$s.

\begin{figure*}
	\centering
	\includegraphics[width=\textwidth]{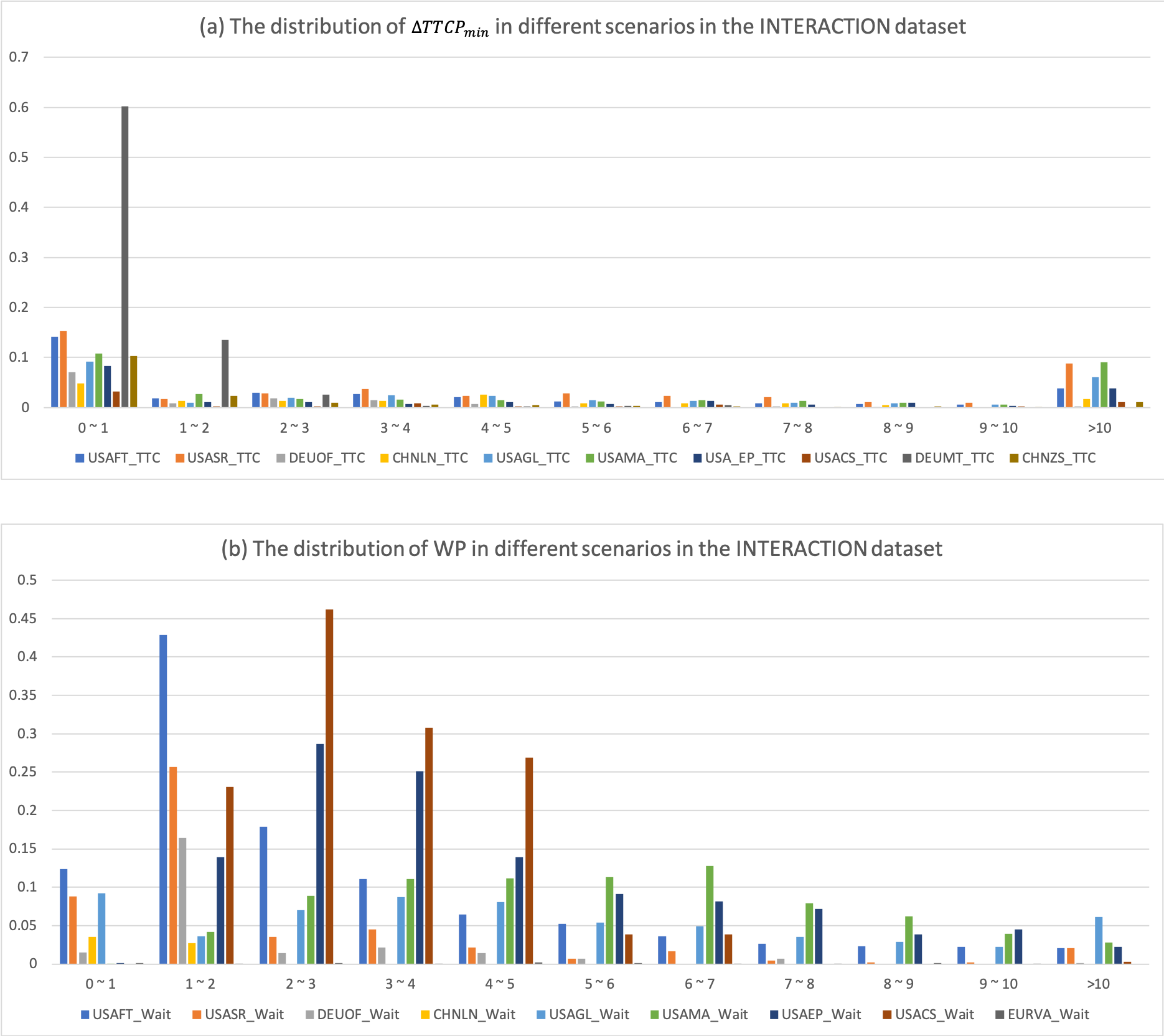}
	\caption{Distribution of the $\triangle TTCP_{\min}$, and WP across different locations and scenarios in the dataset.}
	\label{fig:distribution_statistics_across_scenarios}
\end{figure*}

We also summarized the distributions of $\triangle TTCP_{\min}$ and WP of all vehicles in the dataset over different driving scenarios. The results are shown in \cref{fig:distribution_statistics_across_scenarios}. Similarly, the x-axis represents the length of $\triangle TTCP_{\min}$ and WP in seconds, and the y-axis is the density of vehicles in each scenario. We can see that the dataset contains highly interactive trajectories with a high density of $\triangle TTCP_{\min}\le$\SI{1}{s}, and WP greater than \SI{3}{s}.

\section{Utilization Examples\label{sec: results}}

The proposed dataset is intended to facilitate researches related to driving behavior, as mentioned in Section \ref{sec: introduction}. In this section, we provide several utilization examples of the proposed dataset, including motion/trajectory prediction, imitation learning, motion planning and validation, motion clustering and representation, interaction extraction and human-like behavior generation.

\subsection{Motion Prediction and Behavior Analysis\label{sec: prediction}}

Motion/trajectory prediction is of vital importance for autonomous vehicles, particularly in situations where intensive interaction happens. To obtain an accurate probabilistic prediction model of vehicle motion, both learning- and planning-based approaches have been extensively explored. By providing high-density interactive trajectories along with HD semantic maps, the proposed dataset can be used for both approaches. 

\begin{table}[!htbp]
\begin{center}
\caption{Comparisons of prediction accuracy from \cite{ma_wasserstein_2019}.}
\begin{tabular}{lcccc}
\toprule
\midrule
\textbf{Methods} &  \textbf{features} & \textbf{RMSE} & \textbf{MAE} \\
\midrule
\multirow{3} * {\shortstack[1b]{WAE-based \\approach}}
&$x$	&\textbf{0.013}/\textbf{0.011}		&\textbf{0.046/0.035}	\\
&$y$	&0.006/0.014		&\textbf{0.019/0.041}    \\
&$\psi$	&0.006/\textbf{0.008}						&\textbf{0.018/0.042}   \\
\midrule
\multirow{3} * {\shortstack[1b]{VAE}}
&$x$	&0.018/0.016  		&0.25/0.22  	 \\
&$y$	&\textbf{0.006}/\textbf{0.003}		&0.14/0.22	\\
&$\psi$	&0.006/\textbf{0.008}			&0.13/0.21	\\
\midrule
\multirow{3} * {\shortstack[1b]{Auto-encoder}}
&$x$	&0.315/0.044		&1.026/0.315	\\
&$y$	&0.057/0.141		&0.182/0.479	\\
&$\psi$	&0.011/0.066	&0.037/0.078		\\
\midrule
\multirow{3} * {\shortstack[1b]{GAN}}
&$x$ 		& 0.024/0.020       &0.324/0.273 \\
&$y$ 		& 0.007/0.017		&0.188/0.241\\
&$\psi$	 	& \textbf{0.005}/0.048		&0.107/0.286\\
\bottomrule
\end{tabular}
\end{center}
\label{tab: wae-results}
\end{table}

For instance, \cite{ma_wasserstein_2019} proposed a deep latent variable model based on Wasserstein auto-encoder (WAE) to improve the interpretability. It incorporated the structure of recurrent neural network with vehicle kinematic model such that the output can be constrained. The motion data in \textit{FT} was utilized to train and test the model in comparison with other state-of-the-art models such as variational auto-encoder (VAE), auto-encoder, and generative adversarial network (GAN). Quantitative results shown in \cref{tab: wae-results} demonstrated that the proposed WAE-based method can outperform other state-of-the-art models, when comparing the root mean square error (RMSE) and mean absolute error (MAE) of the prediction for position and yaw angle.

\begin{figure}[htbp]
	\begin{center}
		\includegraphics[width=8.5cm]{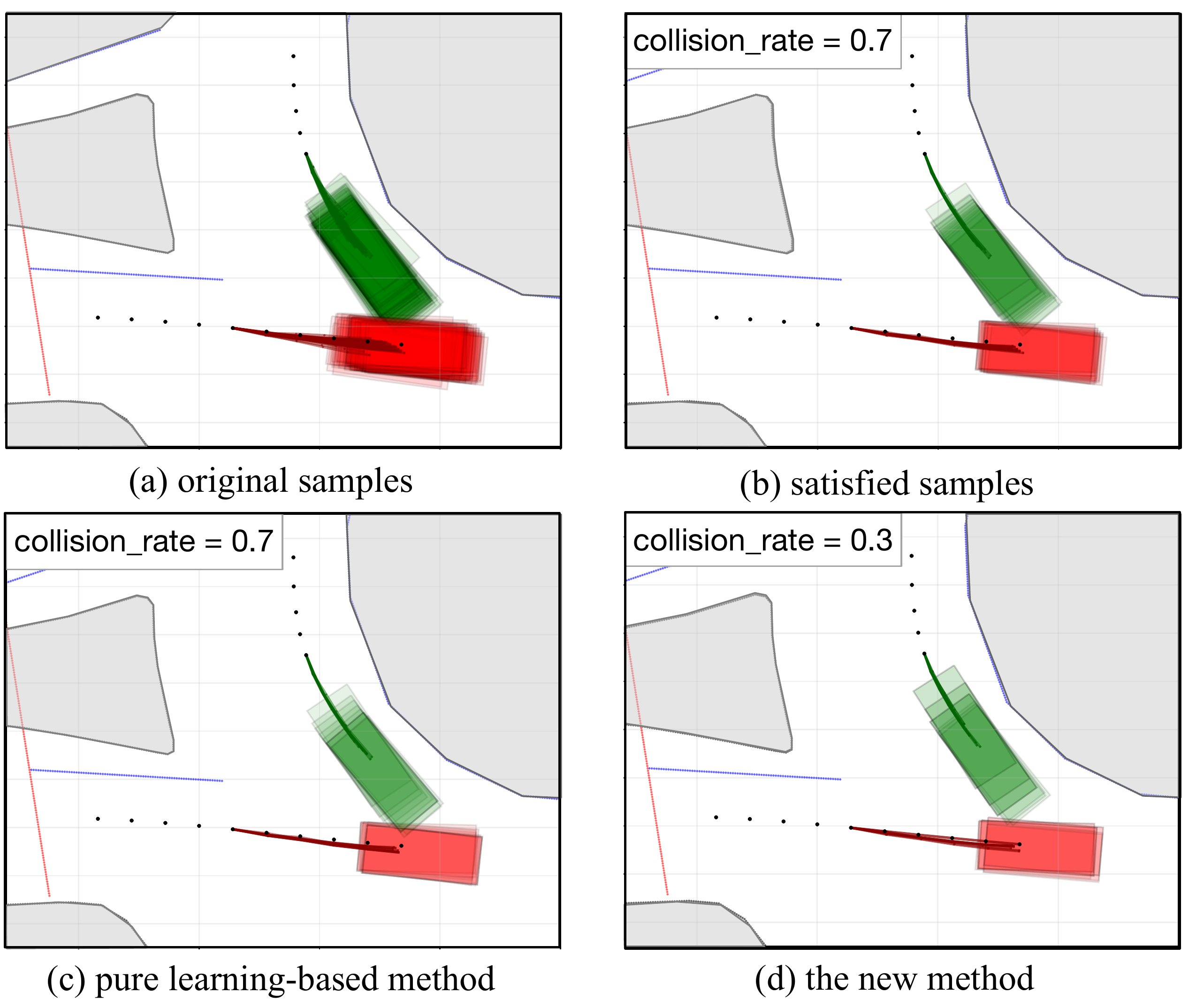}
		\caption{Some exemplar prediction results from \cite{hu2019generic}.}
		\label{fig: yeping2019_results}
	\end{center}
\end{figure}

On the other hand, \cite{hu2019generic} took advantage of the HD semantic maps and combined the learning-based and the planning-based prediction methods. A deep learning model based on conditional variational auto-encoder (CVAE) and an optimal planning framework based on inverse reinforcement learning are dynamically combined to predict both irrational and rational behavior of the vehicles. Benefiting from the the HD semantic information, features for the deep learning model were defined in Frenet frame, which generated much better prediction performance in terms of generalization. Some exemplar results are given in \cref{fig: yeping2019_results}.

\subsection{Imitation Learning\label{sec: imitation}}
\begin{figure}[htbp]
	\centering
	\subfloat[]{\includegraphics[width=0.23\textwidth]{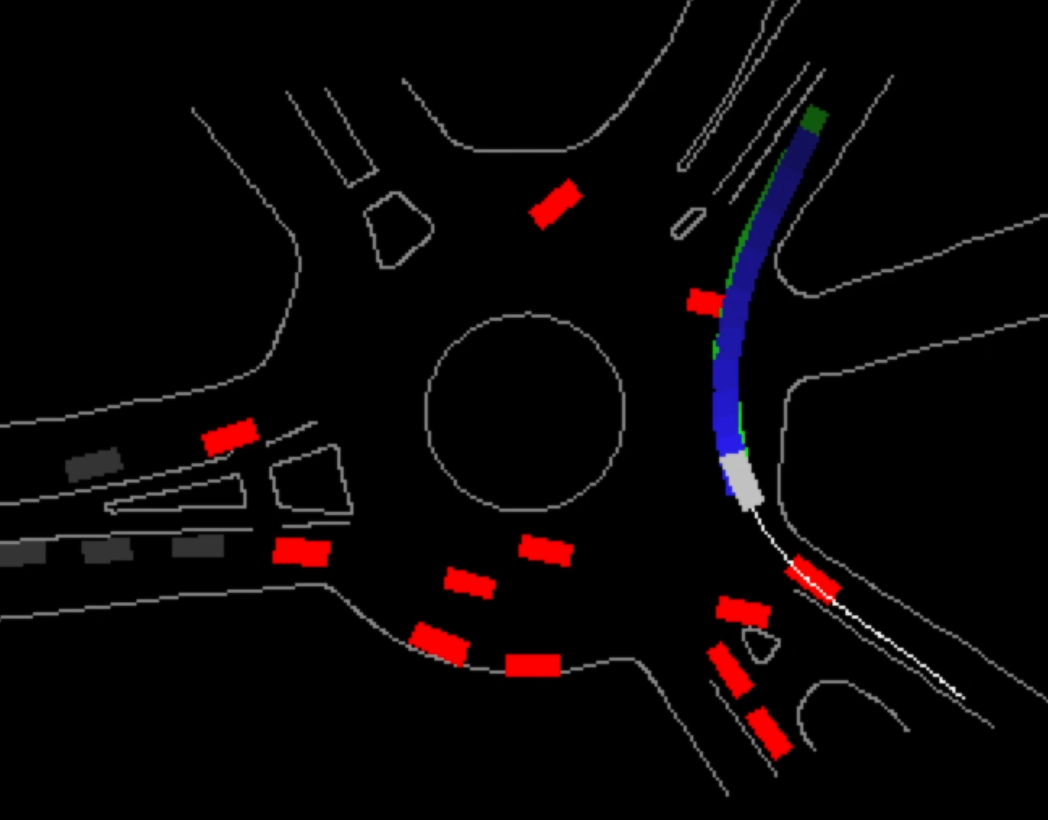}}\quad
	\subfloat[]{\includegraphics[width=0.23\textwidth]{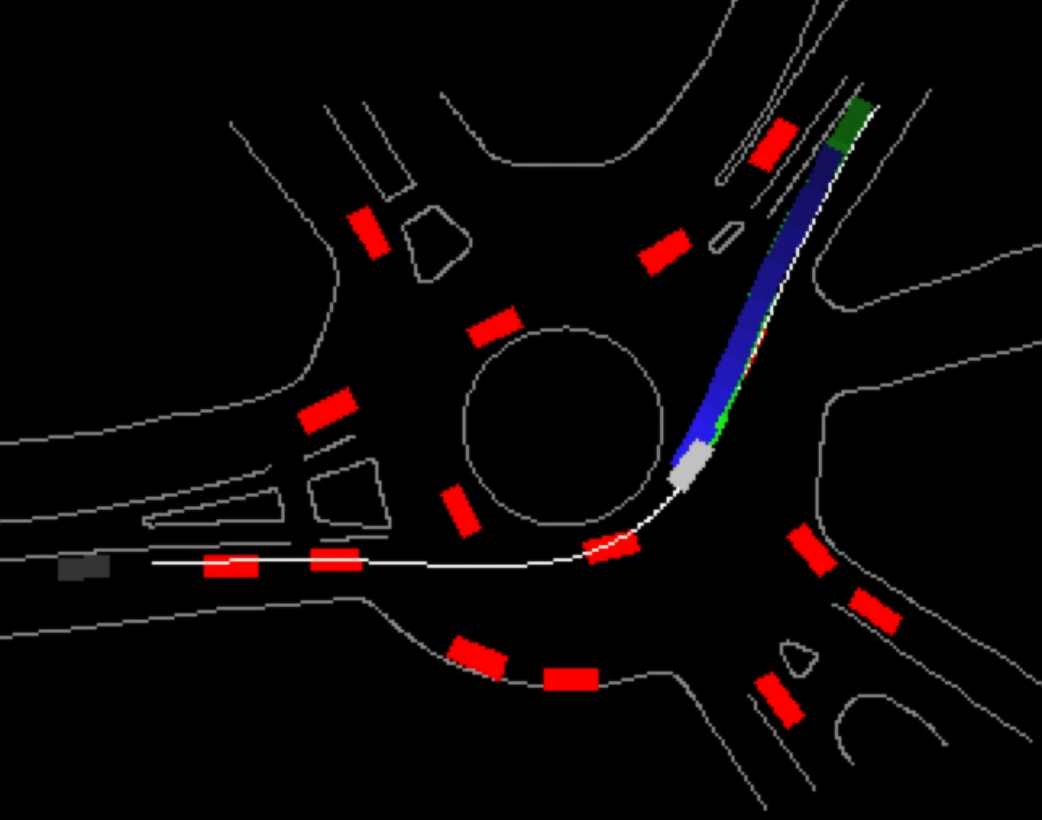}}
	\caption{Two examples of the imitation learning results by employing the method in \cite{sun_fast_2018}.}
	\label{fig:imitation_learning_results}
\end{figure}

The driving behavior in the proposed dataset can also be used for imitation learning which directly imitates how human drive in complicated scenarios. We extended the fast integrated learning and control framework proposed in \cite{sun_fast_2018} in the \textit{FT} roundabout scenario. As shown in \cref{fig:imitation_learning_results}, both the semantic HD map information and the states of surrounding vehicles (the red boxes) were included as the features. The grey box represents the current position of the ego vehicle. The green boxes and blue boxes, respectively, are the ground truth future positions and generated future positions of the ego vehicle via the imitation network.

\subsection{Validation of Decision and Planning}
Besides motion prediction and imitation, the motion data and maps in the dataset can also be used for testing different decision making and motion planning algorithms. The data-replay motions in the dataset are more suitable to test the performances of the decision-maker and planner when the motions of surrounding entities are independent of the ego motions. For example, the motion of the ego vehicle may not effect others when it does not have the RoW, or it has the RoW but others violate the rules or ignore the ego motion.

\begin{figure}[htbp]
	\begin{center}
	\includegraphics[width=8cm]{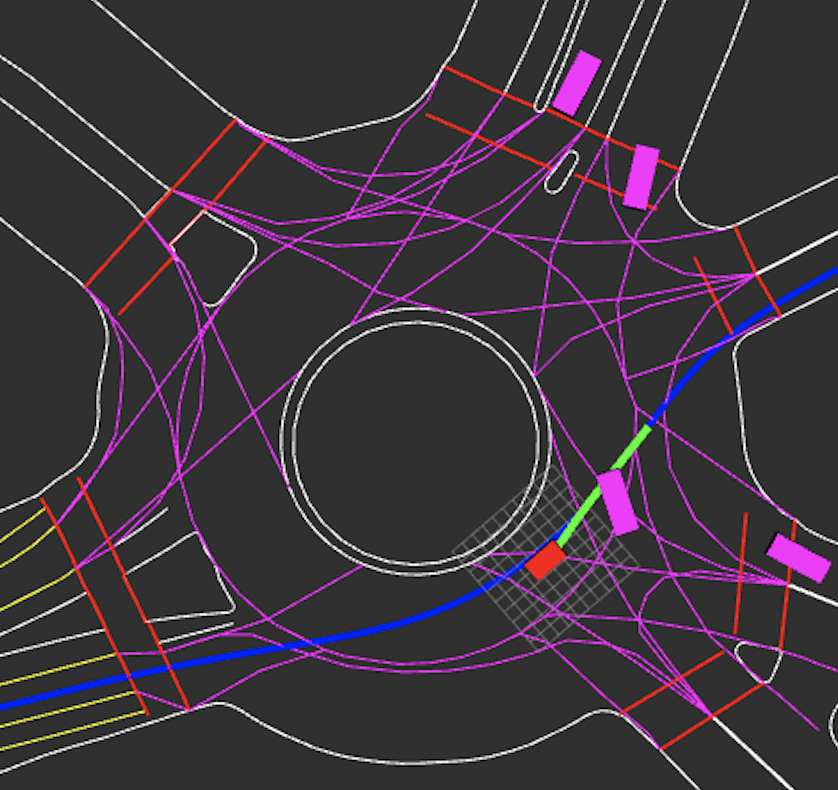}
	\caption{A screen-shot of simulation when testing the motion planner in \cite{zhan_spatially-partitioned_2017} with the proposed dataset.}
	\label{fig: planning-astar-1}
	\end{center}
	\end{figure}

The environmental representation and motion planning methods proposed in \cite{zhan_spatially-partitioned_2017} were tested in the \textit{FT} roundabout scenario. \cref{fig: planning-astar-1} is a bird's-eye-view screen-shot of the simulation. The red rectangle represents the autonomous vehicle with the planner in \cite{zhan_spatially-partitioned_2017}. It was decelerating to avoid the collision with a vehicle entering the roundabout although it has the RoW.
	
We also combined the integrated decision and planning framework proposed in \cite{zhan_non-conservatively_2016} and the sample-based motion planner proposed in \cite{gu2015tunable} to design the decision-maker and planner under uncertainty. The predictor was designed according to \cite{schulz_interaction-aware_2018} based on dynamic Bayesian network (DBN) to provide the probabilities of the intentions of others. 

\begin{figure}[htbp]
	\begin{center}
	\includegraphics[width=8.5cm]{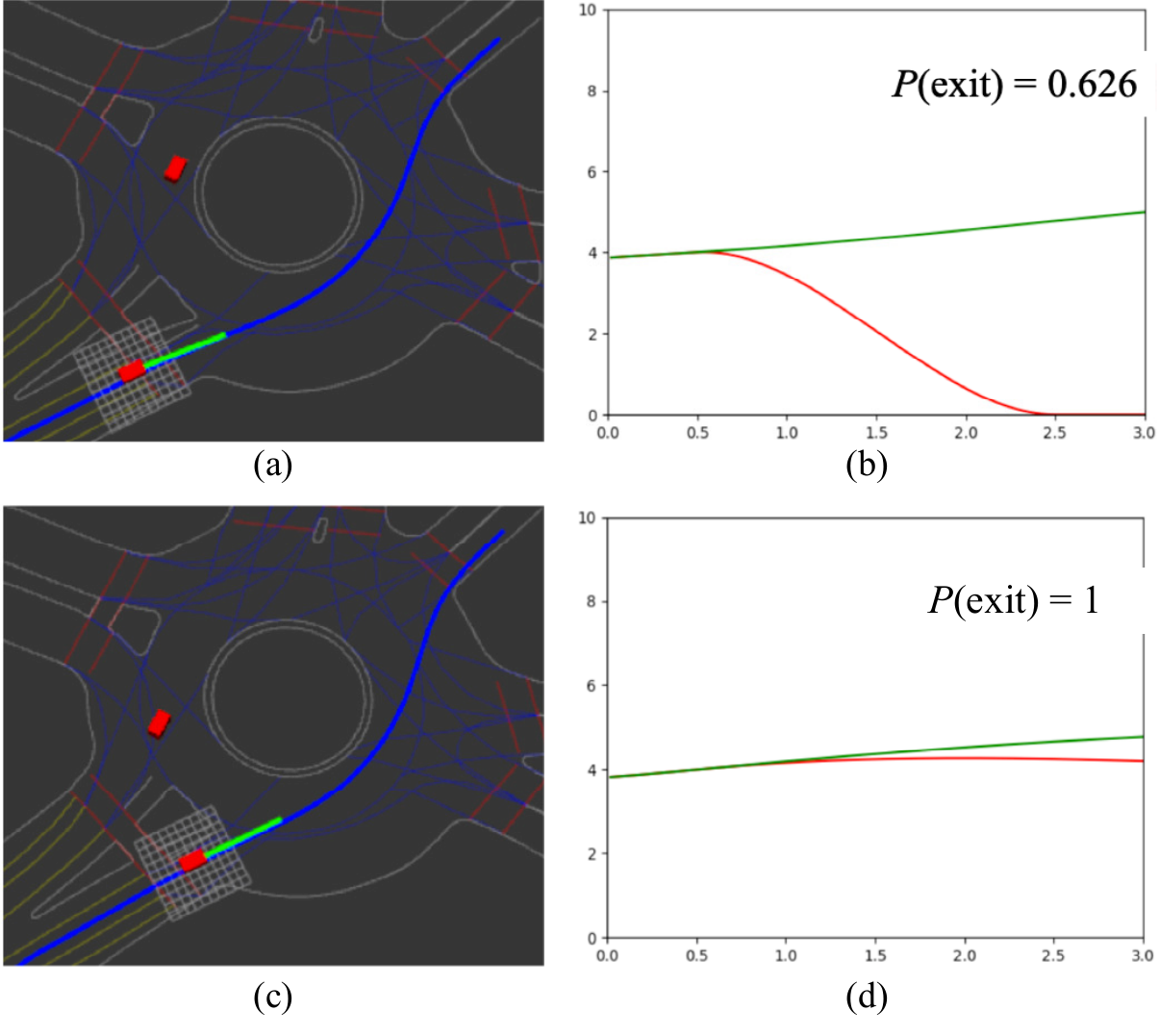}
	\caption{Screenshots of the situations and corresponding planned speed profiles by implementing decision and planning methods in \cite{zhan_non-conservatively_2016, gu2015tunable} with predictor in \cite{schulz_interaction-aware_2018} by utilizing the proposed dataset.}
	\label{fig: planning-ncds}
	\end{center}
	\end{figure}

\Cref{fig: planning-ncds} shows the results of the planned speed profile with corresponding bird's-eye-view screenshots of the situations at specific time steps. The host autonomous vehicle was entering the \textit{FT} roundabout, and the vehicle in the roundabout, retrieved from the proposed dataset, was exiting. When it was not clear whether the target vehicle was going to exit or not, such as the time step in \cref{fig: planning-ncds} (a), the predictor returned $P(\text{exit})$ as $0.626$. With the non-conservatively defensive strategy proposed in \cite{zhan_non-conservatively_2016}, the ego vehicle was able to keep accelerating to enter the roundabout as planned for the next \SI{0.5}{s}, so that the potential threat with low probability (the target vehicle stays in the roundabout) did not affect the efficiency and comfort of the ego vehicle. The long-term planning corresponding to yielding case (red curve in \cref{fig: planning-ncds} (b)) guaranteed that the ego vehicle was able to fully stop for the worst case. 

\subsection{Motion Clustering and Representation Learning\label{sec: clustering}}
\begin{figure}[htbp]
	\begin{center}
	\includegraphics[width=8.8cm]{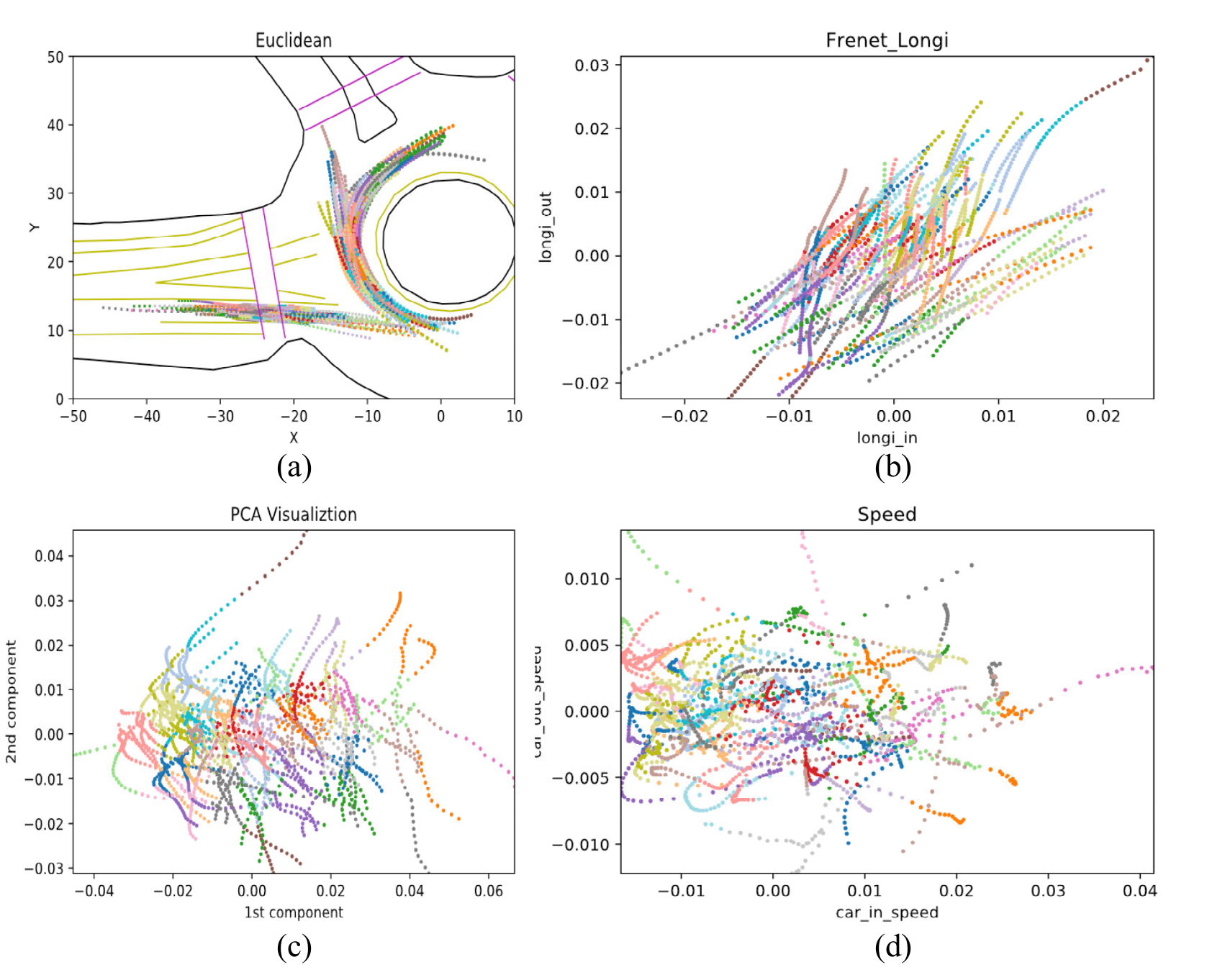}
	\caption{Results of X-means \cite{pelleg2000x} motion clustering using the proposed dataset.}
	\label{fig: xmeans}
	\end{center}
	\end{figure}

The X-means algorithm \cite{pelleg2000x} was employed to cluster the trajectories and obtain motion patterns with results shown in Fig. \ref{fig: xmeans}. We constructed a feature space with vehicle motions in Fren\'et Frame based on map information. Fig. \ref{fig: xmeans} (a) shows the clustered trajectory segments in different colors with the map. Fig. \ref{fig: xmeans} (b) and (d) demonstrate the cluster results with longitudinal positions and speeds of the two interacting vehicles as the coordinates. The clustering results with the first and second components of principle component analysis (PCA) for the feature space are shown in Fig. \ref{fig: xmeans} (c). In the figures we can see that different interactive motions are separated and similar ones are clustered, which are desirable results to obtain motion patterns.

\subsection{Extraction of Interactive Agents and Trajectories\label{sec: extraction}.}
\begin{figure*}[h!]
	\centering
	\subfloat[Paths of two interacting cars]{\includegraphics[width=0.25\textwidth]{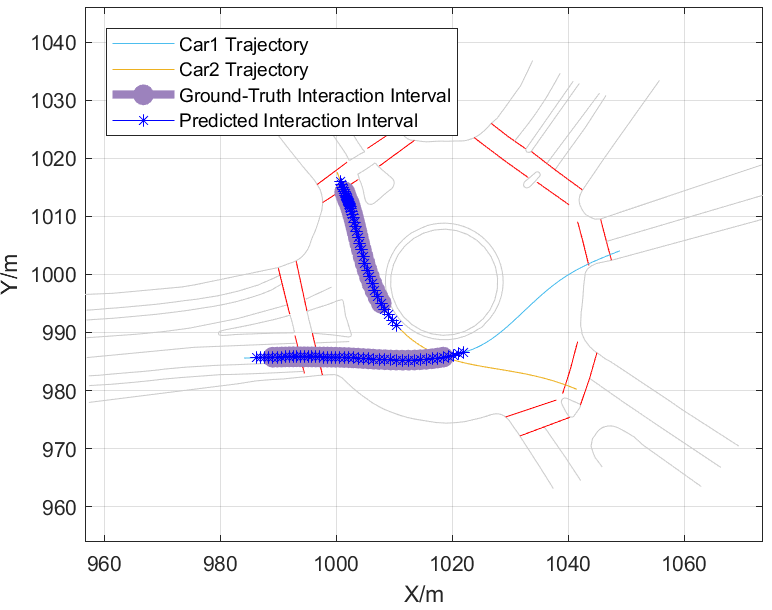}}
	\subfloat[trajectories along longitudinal directions]{\includegraphics[width=0.25\textwidth]{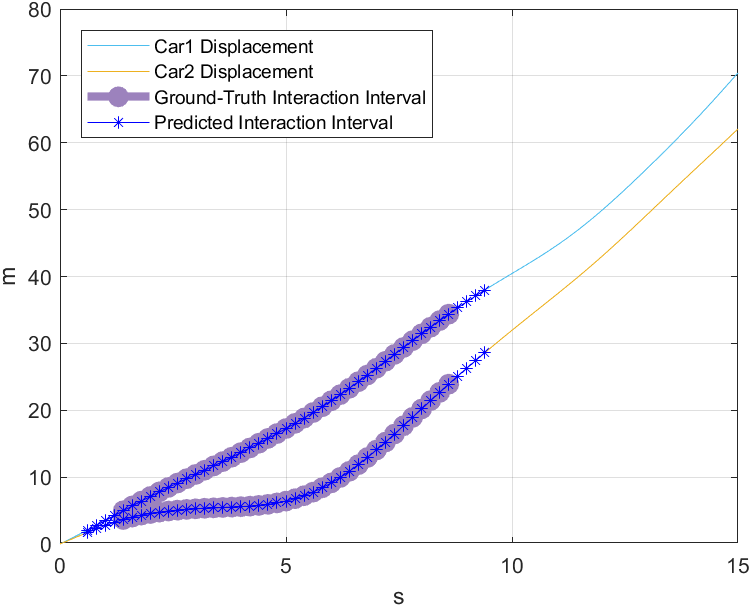}}
	\subfloat[Paths of two interacting cars]{\includegraphics[width=0.25\textwidth]{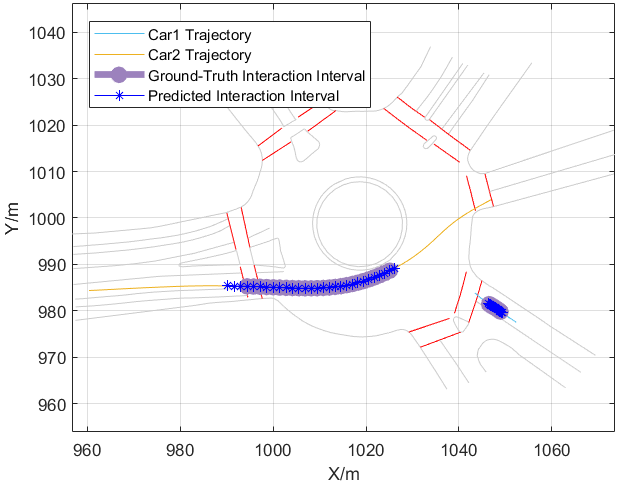}}
	\subfloat[trajectories along longitudinal directions]{\includegraphics[width=0.25\textwidth]{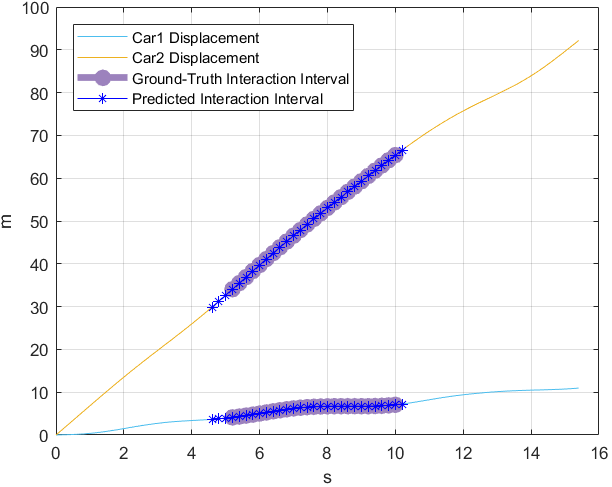}}
	\caption{Two examples of the extracted interaction pairs by implementing the learning method and network structure in \cite{shu2018perception}.}
	\label{fig:example_interaction_extraction}
\end{figure*}

The proposed dataset can also be used to learn the interaction relationships between agents. We implemented the learning method and network structure proposed in \cite{shu2018perception} to extract the interaction frames of two agents. Some example results are given in \cref{fig:example_interaction_extraction}, where \cref{fig:example_interaction_extraction} (a) and (b) provide one exemplar pair of interacting cars in the \textit{FT} scenario, while \cref{fig:example_interaction_extraction} (c) and (d) represent another pair. In \cref{fig:example_interaction_extraction} (a) and (c), the paths of both of the interacting cars are provided, and in \cref{fig:example_interaction_extraction} (b) and (d), the trajectories along longitudinal directions are shown. We can see that the extracted interaction frames (purple circle) align quite well with the ground truth frames (blue star).

\subsection{Human-like Decision and Behavior Generation\label{sec: generation}}

 	\begin{figure}[htbp]
	\begin{center}
		\includegraphics[width=8cm]{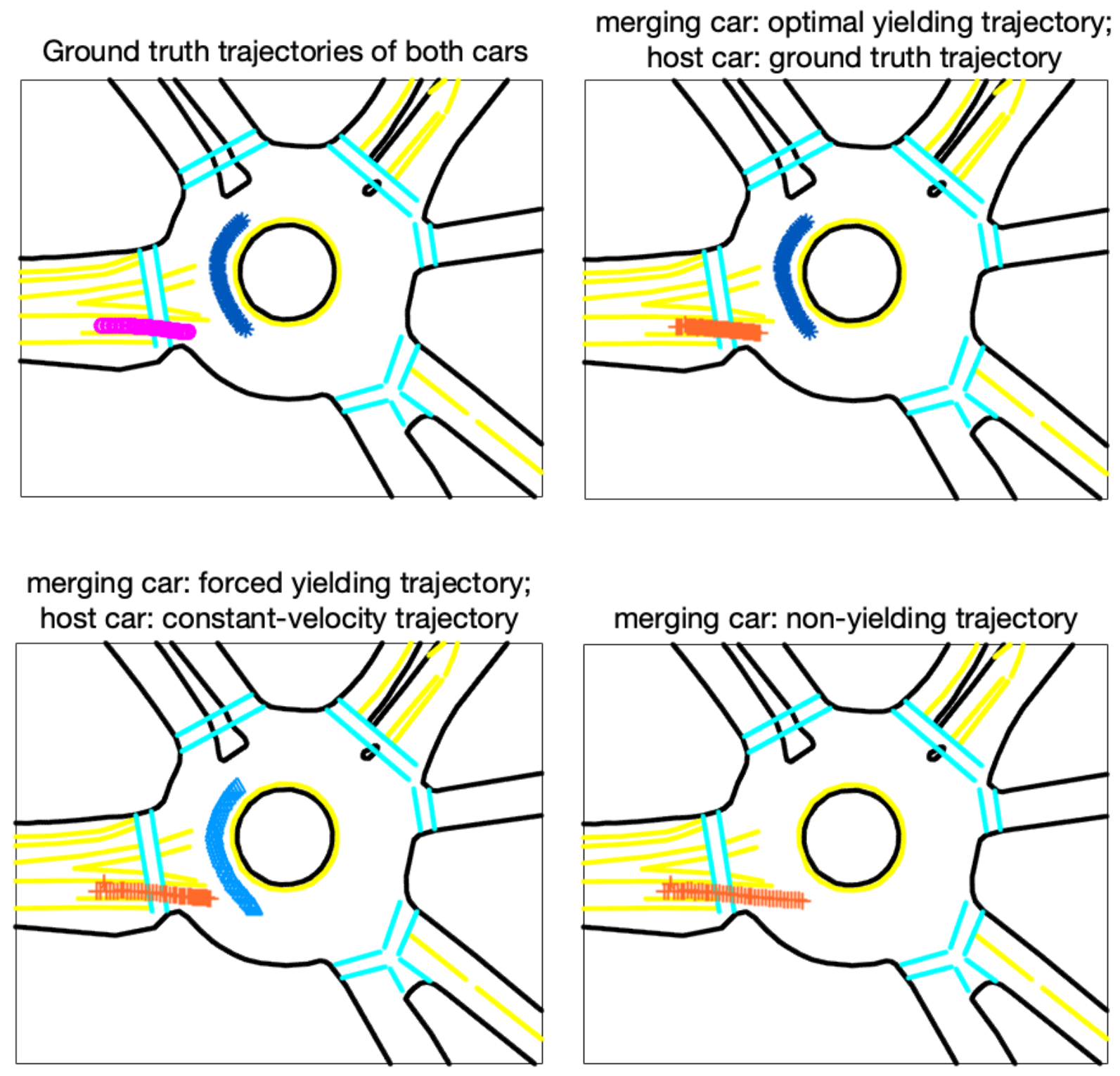}
		\caption{Results of interpretable human behavior model based on the cumulative prospect theory (CPT) \cite{sun_irrational_2019} using the proposed dataset.}
		\label{fig: CPT_result}
	\end{center}
\end{figure}

We can also learn decision-making models that generate human-like decisions and behaviors with the proposed dataset. In \cite{sun_irrational_2019}, an interpretable human behavior model was proposed based on the cumulative prospect theory (CPT). As a non-expected utility theory, CPT can well explain some systematically biased or ``irrational'' behavior/decisions of human that cannot be explained by the expected utility theory. Parameters of three different models were learned and tested using the data in the \textit{FT} roundabout scenario: a predefined model based on time-to-collision-point (TTCP), a learning-based model based on neural networks, and the proposed CPT-based model. The results (\cref{fig: CPT_result}) showed that the CPT-based model outperformed the TTCP model and achieved similar performance as the learning-based model with much less training data and better interpretability.

\section{Conclusion\label{sec: conclusion}}
In this paper, we presented a motion dataset in a variety of highly interactive driving scenarios from the US, Germany, China and other countries, including signalized/unsignalized intersections, roundabouts, ramp merging and lane change from cities and highway. 
Complex interactive motions were captured, featuring inexplicit right-of-way, relatively unstructured roads, as well as aggressive and irrational behavior caused by impatience and social pressure. 
Critical (near-collision and slight-collision) situations can be found in the dataset.
We also included high-definition (HD) maps with semantic information for all scenarios in our dataset. 
The data was recorded from drones and traffic cameras and the data processing pipeline was briefly described.
Our map-aided dataset with diversity, internationality, complexity and criticality of scenarios and behavior can significantly facilitate driving-behavior-related research such as motion prediction, imitation learning, decision-making and planning, representation learning, interaction extraction, and human-like behavior generation, etc. Results from various kinds of methods of these research areas were demonstrated utilizing the proposed dataset.

\section{Acknowledgement\label{sec: acknowledgement}}

The authors also would like to thank the Karlsruhe House of Young Scientists (KHYS) for their support of Maximilian's research visit at MSC Lab.

\bibliography{interactionDataset}

\end{document}